\documentclass[runningheads]{llncs}

\usepackage{eccv}

\usepackage{eccvabbrv}

\usepackage{graphicx}
\usepackage{booktabs}
\usepackage{wrapfig}
\usepackage{setspace}
\usepackage{appendix}
\usepackage{caption}
\usepackage{float}
\usepackage[accsupp]{axessibility}  

\usepackage[pagebackref,breaklinks,colorlinks]{hyperref}

\usepackage{orcidlink}

\newcommand{\tablefont}[1]{\fontsize{#1}{#1}\selectfont}

\begin{document}

\title{HAT: History-Augmented Anchor Transformer for Online Temporal Action Localization} 

\titlerunning{HAT: History-Augmented Anchor Transformer}


\author{Sakib Reza\orcidlink{0000-0001-8491-0316} \and
Yuexi Zhang\orcidlink{0000-0001-5012-5459} \and
Mohsen Moghaddam\orcidlink{0000-0002-3201-6010} \and
Octavia Camps\orcidlink{0000-0003-1945-9172}}

\authorrunning{S. Reza et al.}

\institute{Northeastern University, Boston, MA 02115, USA
\\
\email{\{reza.s,zhang.yuex,m.moghaddam,o.camps\}@northeastern.edu}}

\maketitle

\begin{abstract}
  Online video understanding often relies on  individual frames, leading to frame-by-frame predictions.  Recent advancements such as Online Temporal Action Localization (OnTAL), extend this approach to instance-level predictions. However, existing methods mainly focus on short-term context, neglecting historical information. To address this, we introduce the History-Augmented Anchor Transformer (HAT) Framework for OnTAL. By integrating historical context, our framework enhances the synergy between long-term and short-term information, improving the quality of anchor features crucial for classification and localization. We evaluate our model on both procedural egocentric (PREGO) datasets (EGTEA and EPIC) and standard non-PREGO OnTAL datasets (THUMOS and MUSES). Results show that our model outperforms state-of-the-art approaches significantly on PREGO datasets and achieves comparable or slightly superior performance on non-PREGO datasets, underscoring the importance of leveraging long-term history, especially in procedural and egocentric action scenarios. Code is available at: \url{https://github.com/sakibreza/ECCV24-HAT/}.
\end{abstract}

\section{Introduction}
\label{sec:intro}

In today's digital era, the surge in video content has spurred the demand for innovative solutions to process lengthy, unedited videos. Transformers have become instrumental to understand complex video sequences, particularly in  Temporal Action Localization (TAL) \cite{zhang2022actionformer, cheng2022tallformer, zhao2023re2tal, Shao_2023_ICCV}, which aims to identify and classify actions within videos, marking their start and end times. 

\begin{figure}[t!]
	\centerline{\includegraphics[width=0.8\linewidth]{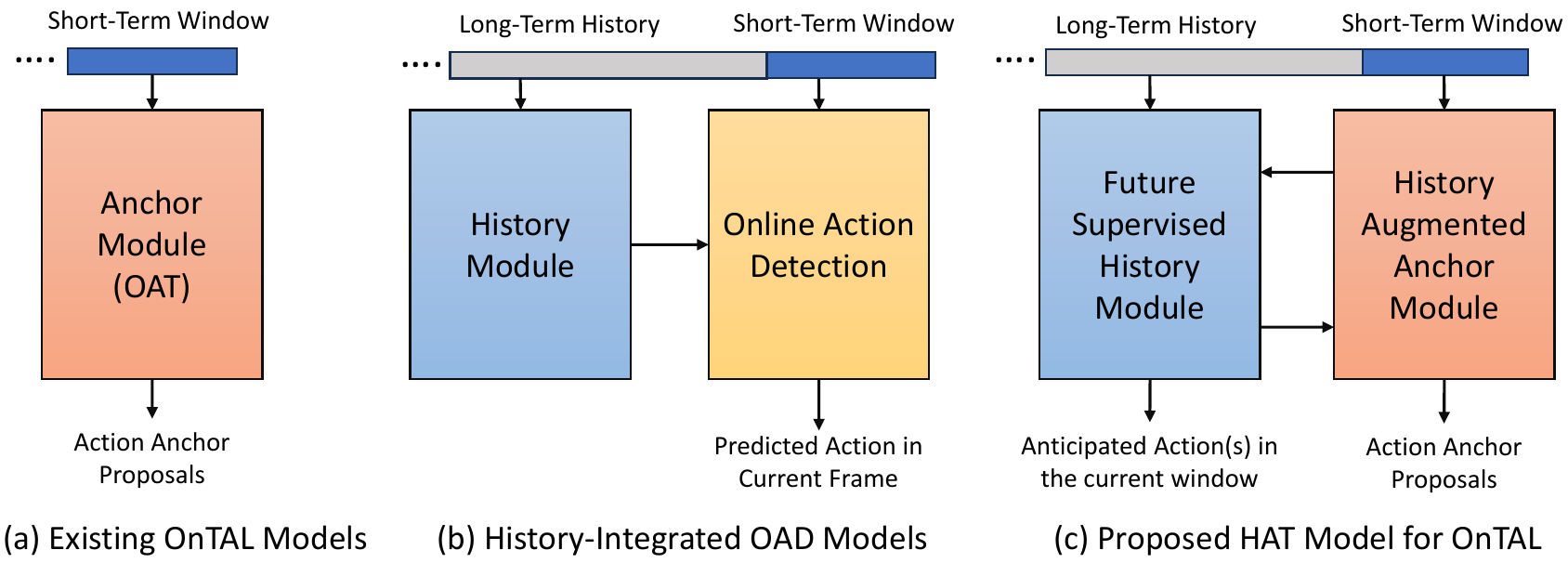}}
 
    \caption{High-level architecture designs of prior approaches and our method.} \label{teaser}
\end{figure}

Online Action Detection (OAD) identifies ongoing actions in real-time video frames, while Online Detection of Action Start (ODAS) focuses on quickly pinpointing action beginnings, enabling prompt responses. However, both tasks operate at the frame level, lacking contextual understanding at the instance level, limiting their broader applicability. To address this, a new instance-level task, Online Temporal Action Localization (On-TAL) \cite{kang2021cag}, has been introduced. On-TAL, unlike traditional offline TAL, operates online without access to future frames and prohibits alterations to previously generated action proposals, allowing only real-time post-processing.

Despite its potential, traditional On-TAL methodologies, which focus on aggregating per-frame action predictions into action instances, encounter challenges in accuracy due to issues like tick, fragmentation, and merging of action instances. To address these, the Online Anchor Transformer (OAT) \cite{kim2022sliding} was introduced, using a Transformer architecture with instance-level supervision. However, OAT's reliance on short-term analysis within a sliding window neglects the potential benefits of incorporating long-term historical context, which is crucial for accurately predicting complex action sequences. While there have been attempts to incorporate historical data in other areas of online action understanding, such as Online Action Detection (OAD)\cite{xu2021long, chen2022gatehub}, the fusion of historical context with OnTAL, particularly with anchor-based Transformer models, remains largely unexplored. Therefore, this paper aims to bridge this gap by integrating historical context into the OnTAL framework, proposing a novel framework named \textbf{H}istory-Augmented \textbf{A}nchor \textbf{T}ransformer (\textbf{HAT}). Figure \ref{teaser} illustrates the overarching structure of our proposed framework in comparison to earlier methods. The main contributions of our paper include:

\begin{itemize}
\item The first long-term historical context supported anchor-based transformer, tailored for OnTAL task, which efficiently processes extensive historical data and integrates it into the anchor-level feature, enhancing the model's understanding and contextual relevance.
\item A novel history refinement strategy that leverages action anticipation guidance and aligns it with the current temporal context to improve prediction accuracy and relevance.
\item A gradient-guided adaptive focal loss function to address the class-imbalance problem effectively, applicable to both foreground versus background and among different foreground classes.
\item A comprehensive evaluation across datasets, showcasing state-of-the-art performance on both procedural egocentric (PREGO) datasets (EGTEA and EPIC-Kitchen-100) and standard non-PREGO OnTAL benchmark datasets (THUMOS'14 and MUSES). Our method particularly stands out in procedural tasks and egocentric datasets due to its advanced understanding of action dependencies and its ability to disregard irrelevant past frames.

\end{itemize}

\section{Related Work}
\label{sec:related}

\paragraph{Temporal Action Localization} TAL detects actions within videos by identifying their start and end points and categorizing them. Initially, researchers used sliding window methods \cite{shou2016temporal, gao2017turn}. Recent studies have expanded the scope of video context analysis, including the use of graph networks \cite{zeng2019graph, xu2020graph, zhao2021videograph}, temporal multi-scale networks \cite{zhao2017temporal, liu2020progressive, zhao2023movement}, multi-scale representation with local self-attention \cite{zhang2022actionformer}, learning based on action sensitivity \cite{Shao_2023_ICCV}, exploring connections among action proposals \cite{chen2019relation}, incorporating overall video context \cite{zhu2021enriching}, and utilizing two-stage long-memory transformers \cite{cheng2022tallformer}. End-to-end training methods \cite{xu2021low, lin2021learning, zhao2023re2tal, kang2023soft, phan2024zeetad} are also gaining interest as an alternative approach in TAL.

\paragraph{Online Video Understanding} 

De Geest et al. \cite{de2016online} pioneered the Online Action Detection (OAD) task, focusing on per-frame action classification in streaming video. Various methods, including reinforcement learning \cite{DBLP:conf/bmvc/GaoYN17a}, future anticipation \cite{xu2019temporal, kim2021temporally, wang2023memory}, GRU-based networks \cite{eun2020learning}, minimal RNNs \cite{an2023miniroad}, and transformer-based approaches \cite{wang2021oadtr, guo2022uncertainty, yang2022colar}, have been utilized for this task. Recent trends favor transformer-based methodologies to leverage historical context for OAD \cite{xu2021long, chen2022gatehub, wang2023memory, guermal2024joadaa}.

Shou et al. \cite{shou2018online} introduced Online Detection of Action Start (ODAS), focusing on identifying action start frames in streaming video. Gao et al. \cite{gao2019startnet} enhanced ODAS with a two-stage approach and policy gradient learning.

Online Temporal Action Localization (On-TAL) emphasizes online detection of action instances, initially proposed by Kang et al. \cite{kang2021cag}. Methods like context-aware actionness grouping (CAG) \cite{kang2021cag}, SimOn \cite{tang2022simon}, 2PESNet \cite{kim20222pesnet}, and OAT model \cite{kim2022sliding} address this task by analyzing streaming videos to generate instance-level action proposals, similar to TAL outputs.

\paragraph{History for Online Action Understanding} 

In online action understanding tasks like OAD, integrating historical context with present information is beneficial. Studies like JOADAA \cite{guermal2024joadaa} use a transformer encoder to process past information and merge it with current features. LSTR \cite{xu2021long} employs transformer decoders for a two-stage compression process to handle extensive historical data efficiently. GateHUB \cite{chen2022gatehub} and MAT \cite{wang2023memory} implement one-stage history compression methods, with GateHUB incorporating a gating mechanism to calibrate attentions and suppress irrelevant frames and backgrounds.

Existing research has explored using long-term historical data in OAD, but similar methods for OnTAL tasks are lacking. To address this, we propose integrating long-term historical context into OnTAL frameworks. However, directly applying techniques from OAD to OnTAL is suboptimal due to task differences: OAD predicts actions at the frame level, while OnTAL identifies discrete action instances.

Thus, we propose a novel history processing and integration method tailored for OnTAL. Our approach compresses long-term historical data efficiently while retaining crucial information for understanding actions at the instance level.

\section{Method}

\subsection{Problem Definition}
In this study, we tackle the Online Temporal Action Localization (On-TAL) problem \cite{kang2021cag}, where the task is to identify and classify action instances within an untrimmed video stream, \(V = \{v_i\}_{i=1}^T\), consisting of \(T\) frames, alongside its corresponding \(K\) label sets, \(\Psi = \{\psi_k\}_{k=1}^K = \{(s_k, e_k, c_k)\}_{k=1}^K\), where \(s_k\), \(e_k\), and \(c_k\) denote the start time, end time, and action category of the \(k^{th}\) action instance, respectively. The inherent online constraint of On-TAL stipulates that at any given timestamp \(t\) (\(1 \leq t \leq T\)), only the partial video \(V_{1:t} = \{v_i\}_{i=1}^t\) is observable, necessitating the generation of action proposals immediately upon detection of action ends, with the ultimate goal of sequentially recovering \(\Psi\). This paradigm imposes a stringent condition that once an action proposal is generated, it cannot be subsequently revised or removed.


\subsection{Proposed Framework Overview}

\begin{figure}[h!]
	\centerline{\includegraphics[width=1\linewidth]{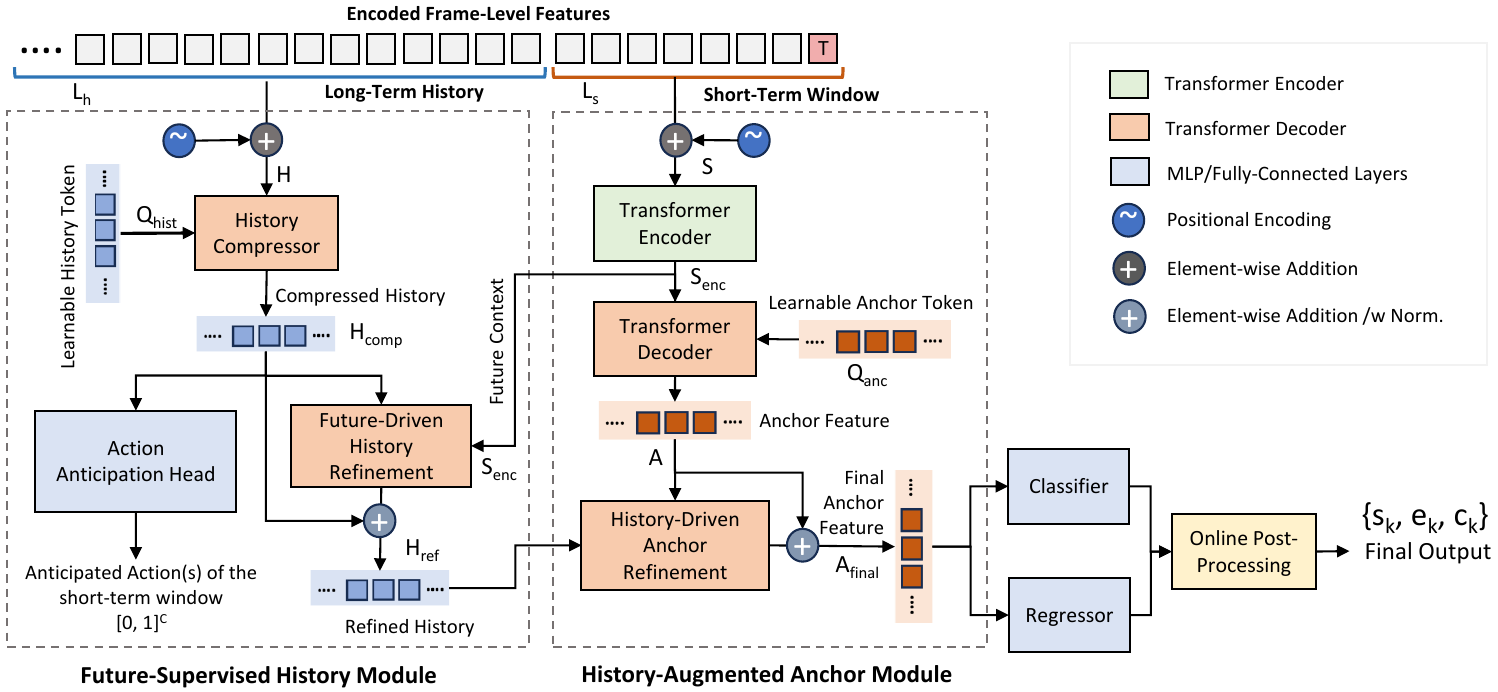}}
    \caption{Proposed History-Augmented Anchor Transformer (HAT) architecture.} \label{model}
\end{figure}
Figure \ref{model} illustrates the architecture of the proposed History-Augmented Anchor Transformer (HAT) framework. 
Its {\bf history module} handles long-term  context using a compressor. The compressed history is passed to an action anticipation head, which utilizes window-level supervision for current action anticipation to calibrate the compressor. Simultaneously, the compressed historical data is processed through future-driven history refinement to ensure alignment with the current context. The {\bf history-augmented anchor module} processes short-term window features, utilizing a transformer encoder to capture temporal context and a transformer decoder to generate anchor features. These features,  enhanced with the history information, are the augmented anchor features used in the {\bf prediction module}. This module contains a classifier and a regressor to generate action proposals for anchors, followed by an online post-processing step to eliminate redundant proposals. A detailed description is given next, with additional information in the appendix \ref{appx:method}.

\subsection{Initial Feature Extraction}

Each input video segment  undergoes an initial transformation into a  frame-level encoded representation using a pre-trained feature encoder (e.g., I3D\cite{i3d}, SlowFast\cite{feichtenhofer2019slowfast}). At a given time point $t$, these vectors are mapped into a D-dimensional feature space via a linear projection layer, represented as \( F = \{ f_i \}_{i=t-L_q+1}^{t} \), where \(L_q\) is the total length of the input queue. The feature $F$ is further divided into two distinct subsets to capture different temporal contexts: the long-term history, denoted as \( H = \{ PE(f_i) \}_{i=t-L_q+1}^{t-L_s} \), and the short-term temporal window, denoted as \( S = \{ PE(f_i) \}_{i=t-L_s+1}^{t} \), where $PE$ is the positional encoding function, $L_s$ and $L_h= L_q - L_s$ are the lengths of the short-term and long-term history windows, respectively.  The long-term component, \(H \in \mathbb{R}^{L_h \times D}\), is processed further in the history module, while the short-term segment, \(S \in \mathbb{R}^{L_s \times D} \), is utilized within the anchor module to produce anchor features.

\subsection{Future-Supervised History Module}
The purpose of the future-supervised history module is to compress the long-term historical data (H) to enhance processing efficiency while preserving vital historical information. This module further refines the compressed data by incorporating weak anticipatory supervision of actions and alignment with future contexts. The methodology is detailed next.\\


\noindent {\bf History Compressor.} The  long-term history \(L_h\) can be significantly long. Utilizing self-attention mechanisms to encapsulate historical temporal information is computationally intensive as the complexity scales quadratically with \(L_h\). Moreover, integrating this information with anchor features further increases computational demands. Inspired by prior work \cite{xu2021long}, we introduce a history compression strategy to mitigate these challenges. This method employs a standard transformer decoder integrated with a learnable historical token, \(Q_{\text{hist}}\), of size \(L_{\text{comp}}\), where \(L_{\text{comp}} << L_h\). The compression is done through  transformer decoder blocks performing the following operations: 
\begin{align} 
d(Q, X) &= \text{Norm}(X_{\text{attn}} + \text{FFN}(X_{\text{attn}})), \label{eq:1} \\  
X_{\text{attn}} &= \text{Norm}(Q_{\text{attn}} + \text{CrossAttn}(Q_{\text{attn}}, X)), \\
Q_{\text{attn}} &= \text{SelfAttn}(Q),  
\end{align} 
where  the decoder \(d\) integrates  multi-head self-attention (SelfAttn), multi-head cross-attention (CrossAttn), layer normalization (Norm), and a feed-forward network (FFN). For this compression operation, we set the query tokens \(Q=Q_\text{hist}\), with the learnable history token, and the input token \(X = H\), with the initial history feature. The architecture incorporates \(N_c\) identical decoder blocks, to generate the compressed history, \( H_\text{comp} = d_{N_c}  \circ \ldots \circ d_1 (Q_\text{hist}, H) \in \mathbb{R}^{L_\text{comp} \times D} \).\\

\noindent {\bf Action Anticipation Head.}
To enhance the efficacy of the history compression, it is crucial to retain information from frames that play a significant role in predicting ongoing action instances. In this light, we introduce an innovative concept: the action anticipation head, which processes the compressed history and predicts actions within the current sliding window \( S \) (considered as future actions by the history encoder). Our method prioritizes prediction at the window level, rather than at the frame-wise or anchor-level, thereby providing  weak supervision through forward anticipation. This novel approach enables the history compressor to calibrate its attention by predicting future actions, allocating increased attention to frames that are pivotal in predicting action instances while diminishing emphasis on less relevant frames.

In our architecture, the condensed historical feature \( H_\text{comp} \in \mathbb{R}^{L_\text{comp} \times D} \) is fed into the action anticipation head to predict forthcoming actions. The anticipation mechanism is built around a simple neural network structure. Initially, a linear layer reduces the feature dimension from \( D \) to \( \frac{D}{4} \). Subsequently, the features are flattened to prepare them for sequential processing. Finally, two densely connected layers are employed, with the first utilizing the ReLU activation function and the last utilizing the sigmoid function. This architecture results in the generation of predicted action probabilities within the current sliding window \( S \), denoted as \( a \in [0, 1]^{1 \times C} \), where C is the number of action classes.

\subsubsection
\noindent {\bf Future-Driven History Refinement.}
The objective of the future-driven history refinement module is to enhance the intermediate history feature by aligning it with the short-term temporal context, thereby making it more pertinent in understanding current action instances. Its core  is built around a transformer decoder (Equation \ref{eq:1}) that takes two inputs: the compressed history (\(H_\text{comp}\)) and the short-term temporal features (\(S_\text{enc}\)). This process is executed through a series of transformations with \(N_r\) blocks, depicted as \( H'_\text{comp} = d_{N_r} \circ \ldots \circ d_1 (H_\text{comp}, S_\text{enc}) \in \mathbb{R}^{L_\text{comp} \times D} \). The cross-attention mechanism in the transformer decoder uses the current short-term context to calibrate the attention weights, ensuring that historical features relevant to the current context are emphasized, while others are de-emphasized.  Additionally, to ensure the preservation of the intermediate historical context, we add this refined feature back to the initial compressed history and then apply normalization, resulting in the final refined history, \( H_\text{ref} = \text{Norm}(H'_\text{comp} + H_\text{comp}) \). This method allows us to enhance historical data with short-term context while retaining the intermediate historical information.

\subsection{History-Augmented Anchor Module}

\subsubsection{Anchor Feature Generation}
In our framework,  we follow the approach used in OAT \cite{kim2022sliding} to generate anchor features. It involves two main stages. First, a transformer encoder, composed of \(N_e\) sequential blocks, is utilized to convert local attributes into a temporal contextual feature, \(S_\text{enc}\  \in \mathbb{R}^{L_s \times D}\), which is then supplied to a transformer decoder. This decoder also receives as input a learnable anchor query \(Q_\text{anc} \in \mathbb{R}^{M \times D}\), where M is the number of pre-defined anchors. As data progresses through its \(N_d\)  blocks, the decoder produces the anchor feature A \( \in \mathbb{R}^{M \times D}\), where each of these \(M\) feature vectors represents a pre-defined anchor.

\subsubsection{History-Driven Anchor Refinement}
In this module, we augment the initial anchor features by integrating them with refined historical data, thereby providing a more comprehensive context. To combine two sets of features, rather than using standard approaches such as concatenation and average pooling, we explore more capable techniques employing attention mechanisms. Specifically, cross-attention stands out as a better way to integrate features from different domains, particularly when integrating past context with the current \cite{xu2021long}. Here, we utilize a standard Transformer Decoder, consisting of \(N_a\) decoder blocks (represented by \(d\) in equation \ref{eq:1}), to handle the initial anchor feature, \(A\), and the refined historical context, \(H_\text{ref}\), obtained from our history module. Additionally, we add the enhanced anchor feature with the original anchor feature and apply normalization to derive the final anchor feature representation \( A_\text{final} \in \mathbb{R}^{M \times D} \). This approach ensures that the model retains original anchor information while augmenting it with historical context. The resulting final anchor feature is subsequently used in both classification and regression tasks to generate action proposals for the pre-defined anchors.

\subsection{Prediction Module and Online Post-Processing}
In our framework, we adopt the prediction module and online post-processing techniques as outlined in our baseline work \cite{kim2022sliding}. The prediction module consists of an action classifier and a boundary regressor. Both components use the same processed anchor feature representation \(A\) for their input. The function of the action classifier is to identify the specific action (or background) taking place within a given anchor segment. Conversely, the regressor is designed to compute two key measurements: the offset between the end of the target action and the anchor segment's end, and the ratio of the duration of the target action to the total duration of the anchor segment. In each frame, multiple action proposals are generated and filtered using Non-Maximum Suppression (NMS). The probabilities of the selected proposals are then queued and inputted into the Online Suppression Network (OSN), which determines their inclusion in the final instance set, \(\Psi\). OSN functions as a substitute for the traditional offline NMS method, adhering to online constraints.

\subsection{Training Strategy and Loss Function}
\noindent {\bf Adaptive Focal Loss.}
In the video action understanding datasets, a common issue is the imbalance between action samples and background samples, with the latter dominating. Chen et. al. \cite{chen2022gatehub} proposed a modified focal loss to solve this imbalance by assigning varying levels of emphasis to different instances based on their classification confidence. Specifically, it differentiates between hard-to-classify action instances and background instances by applying distinct focal factors, thereby addressing the imbalance between action and background classes. However, this approach does not address the potential imbalance among foreground action classes. In response, inspired by equalized focal loss used in object detection tasks \cite{li2022equalized}, we introduce a new loss function for OnTAL, named Adaptive Focal Loss (AFL). This new loss function dynamically adjusts the focal factor for each action class during training, informed by the gradients, to better manage class imbalance across the foreground action categories. The Adaptive Focal Loss is formulated as follows:
\begin{align}
AFL(p_i, y_i) = \sum_{j=0}^{C} -y_i^j (1 - p_i^j)^{\lambda^j} \log(p_i^j),
\end{align}
where \( p_i \) is the predicted probability for the \(i_{}^{th}\) anchor, \( y_i \) is the corresponding ground truth label, and \( \lambda^j \) is the class-specific focal factor for the \(j_{}^{th}\) action category. This class-specific focal factor, \( \lambda^j \), is defined as $\lambda^j =  \lambda_b$ for $j=C$ (background), and $\lambda^j =\lambda_b + \lambda^j_f$ for foreground classes,
where \( \lambda_b \) is the base focal factor, and \( \lambda^j_f = s(1 - r^j) \) is an adaptive component for the foreground classes, with \( r^j \) being the ratio of accumulated gradients for positive versus negative samples for the j-th category. The scaling factor \( s \) adjusts the influence of \( \lambda^j_f \). The value of \( r^j \) approaches 1 for categories with less imbalance, thereby reducing the additional focal impact, and approaches 0 for categories with greater imbalance, thereby increasing the additional focal impact. Further details on this loss are available in the appendix \ref{appx:training}.\\

\noindent {\bf Final Loss.} 
In the training phase, we compute the training objective similarly to the established baseline \cite{kim2022sliding}, with the exception that we apply our newly introduced adaptive focal loss (AFL) for classification tasks. To determine the overall loss for the framework, we utilize the AFL to compute the classification loss, \(L_c\), alongside the regression losses for offset, \(L_o\), and length adjustment, \(L_l\), which are both calculated using L1 loss. Moreover, for the action anticipation head, we introduce an additional classification loss using AFL, denoted as \(L_a\). Consequently, the overall loss is formulated as \(\mathcal{L} = \alpha \mathcal{L}_c + \beta (\mathcal{L}_o + \mathcal{L}_l) + \gamma \mathcal{L}_a\), where \(\alpha\), \(\beta\), and \(\gamma\) represent the weighted coefficients assigned to the classification, regression, and action anticipation losses, respectively. The training and evaluation procedures are carried out in a manner similar to the baseline, with the key difference being that during inference, the output from the action anticipation head is disregarded since it serves solely as a training enhancement.

\section{Experiments}

\subsection{Datasets}

Commonly used datasets  like THUMOS and MUSES lack action variety within each video, limiting their effectiveness in testing OnTAL models. To overcome this shortfall, we incorporate two egocentric procedural action datasets, EGTEA and EPIC-Kitchen-100 (EK100) to our testing. These datasets provide longer videos, a broader range of action categories, and a higher frequency of different actions per video. Importantly, the procedural nature of these actions ensures they are contextually linked to previous ones, offering a solid basis for evaluating the OnTAL models' ability to integrate historical context. Additionally, the unique egocentric viewpoint leads to situations where actions occur outside the camera's view or are obscured by the user's hand, requiring the model to deduce actions based on temporal relationships. These characteristics make EGTEA and EK100 perfect for a more comprehensive and realistic evaluation of model performance in dynamic, real-life environments. 
Thus, we focus on EGTEA and EK100 for validating our proposed framework, while also presenting results for THUMOS and MUSES.

EGTEA \cite{li2018eye} includes 28 hours of egocentric video content, comprising 86 sessions from 32 subjects performing  kitchen tasks. It features 22 different action classes, with an average of 180 action instances and 15.6 action classes per video.

EPIC-Kitchen \cite{damen2022rescaling} is one of the largest egocentric datasets available, containing 100 hours of video from 700 sessions. These videos capture various cooking activities in different kitchens, with an average of 121.5 action instances and 14.5 action classes per video.

THUMOS’14 \cite{idrees2017thumos} consists of 200 training videos and 213 test videos, covering 20 action classes related to sports. These are untrimmed videos with an average of 15.5 action instances and only 1.2 action classes per video.

MUSES \cite{liu2021multi} comprises 3,697 videos, amounting to a total of 716 hours, and featuring 25 action classes. These videos are taken from TV and movie dramas, with an average of 8.5 action instances and 3.3 action classes per video. 

\subsection{Implementation Details}
In our initial extraction of video features for EGTEA, we use the I3D model pre-trained on the Kinetics dataset\cite{i3d}. The video frames were processed at their original frame rate of 24 FPS and a stride of 12, yielding two feature vectors per second. For the EPIC-Kitchens dataset, we adopted SlowFast features in line with the settings utilized by state-of-the-art TAL literature. For the THUMOS’14 dataset, we used the two-stream TSN approach \cite{wang2016temporal}, which was trained on the Kinetics dataset, following the methodology of prior research. Lastly, for the MUSES dataset, we used the I3D feature extractor again, aligning with the standards established by previous studies on MUSES.

In our framework configuration, we utilize a transformer with D = 1024 feature dimensions. In our architecture, transformer decoders have \(N_c = N_d = N_a = 5\) blocks with 4 heads each, except the future-driven refinement decoder with \(N_r = 2\) blocks with 4 heads each. The transformer encoder has \(N_e = 3\) blocks with 8 heads. Regarding dataset-specific parameters, for the EGTEA and EPIC-Kitchens datasets, we set the history length \(L_h\) = 48 and the short-term window length \(L_s\) = 16, which respectively correspond to 24 seconds of historical data and 8 seconds of short-term data. Additionally, we employ \(M=6\) anchor sizes: {2, 4, 6, 8, 12, 16}. For the THUMOS’14 dataset, the history and short-term lengths are set to \(L_h = 256\) and \(L_s = 64\), with \(M=6\) anchor sizes of {4, 8, 16, 32, 48, 64}. For the MUSES dataset, we set \(L_h = 300\), \(L_s = 150\), and use \(M=7\) anchor sizes: {4, 9, 18, 37, 75, 112, 150}. In the adaptive focal loss function, we define the parameters as follows: \(\lambda_b = 0.025\) and \(s = 0.05\). Additionally, the weighted coefficients for the final loss are specified as \(\alpha = \beta = 1\) and \(\gamma = 0.2\). For other hyperparameters and the training of both the main and suppression networks, we adhere to the settings specified in OAT \cite{kim2022sliding}. Additionally, our model operates under the online execution setting as mentioned in \cite{kim2022sliding}, achieving an inference speed of 147 FPS on an Nvidia GeForce RTX 4090 GPU.

\subsection{Comparison with the State of the Art}
Since OnTAL is a relatively new and less investigated area, there are only a few existing models available for comparison. Our model is primarily evaluated against the current leading OnTAL model, OAT. Additionally, we have included the performance metrics of other available models for comparison, as reported in the existing literature.
\begin{table}
\tablefont{2.0mm} 
\singlespacing
\caption{Comparison of mAP (\%) across various tIOU thresholds with OnTAL baseline on the Procedural Egocentric (PREGO) Datasets. }
\label{Tab:1}

  \centering

  \begin{tabular}{l l c c c c c c} 
 \toprule
  Dataset & Model &  0.1 & 0.2 & 0.3 & 0.4 & 0.5 & Avg. \\ [0.5ex] 
 \hline
     EGTEA & OAT \cite{kim2022sliding} & 24.9 & 23.1 & 20.6 & 17.1 & 12.2 & 19.6 \\
     & HAT (Ours) & \textbf{27.5} & \textbf{25.4} & \textbf{22.6} & \textbf{18.3} & \textbf{13.5} & \textbf{21.5} \\
 \hline
     EK-100 & OAT \cite{kim2022sliding} & 17.8 & 16.3 & 14.3 & 12.8 & 10.1 & 14.2 \\
     & HAT (Ours) & \textbf{18.3} & \textbf{17.0} & \textbf{15.8} & \textbf{13.9} & \textbf{11.5} & \textbf{15.3} \\
    
\bottomrule
\end{tabular}

\end{table}

In Table \ref{Tab:1}, we present the performance of our model on the Procedural Egocentric (PREGO) datasets, specifically the EGTEA and EPIC-KITCHENS datasets. Our findings indicate a notable improvement over the current state-of-the-art OnTAL model, OAT. This enhancement underlines the effectiveness of our proposed history integration technique in understanding procedural and egocentric instance-level actions. The results convincingly demonstrate that our model holds a significant advantage in handling PREGO scenarios.

\begin{table}
\tablefont{2.0mm} 
\singlespacing
\caption{Comparison of mAP (\%) across various tIOU thresholds on the Non-PREGO Standard OnTAL Datasets. (*Reconstructed result)}
\label{Tab:2}
  \centering

  \begin{tabular}{l l c c c c c c} 
 \toprule
  Dataset & Model &  0.3 & 0.4 & 0.5 & 0.6 & 0.7 & Avg. \\ [0.5ex] 
 \hline
     THUMOS'14   & CAG-QIL\cite{kang2021cag} & 44.7 & 37.6 & 29.8 & 21.9 & 14.5 & 29.7
    \\
    & 2PESNet\cite{kim20222pesnet} & 47.4 & 39.8 & 31.4 & 21.8 & 14.0 & 30.9
    \\
    & SimOn\cite{tang2022simon} & 57.0 & 47.5 & 37.3 & 26.6 & 16.0 & 36.9
    \\
     & OAT\cite{kim2022sliding} & \textbf{63.0} & 56.7 & 47.1 & 36.3 & 20.0 & 44.6 \\
     & HAT (Ours) & 62.0 & \textbf{57.0} & \textbf{48.0} & \textbf{36.5} & \textbf{20.7} & \textbf{44.8} \\
 \hline
    MUSES & CAG-QIL\cite{kang2021cag} & 8.5 & 6.5 & 4.2 & 2.8 & 1.9 & 4.8 \\

     & OAT*\cite{kim2022sliding} & 16.7 & 13.0 & 10.0 & \textbf{6.3} & 3.2 & 9.8 \\
     & HAT (Ours) & \textbf{19.1} & \textbf{14.7} & \textbf{10.1} & \textbf{6.3} & \textbf{3.7} & \textbf{10.8} \\
    
\bottomrule
\end{tabular}

\end{table}

On the other hand, Table \ref{Tab:2} shows our model's performance on non-PREGO datasets, namely THUMOS and MUSES. Here, our model displays performance levels that are comparable to, or in some instances slightly superior to, existing models (with some exceptions). These results suggest that while our model is still effective in non-PREGO contexts, the impact of historical information is more pronounced and beneficial in PREGO situations. This distinction underscores the specialized advantage of our model in scenarios requiring nuanced, instance-level action understanding within procedural and egocentric contexts.

\begin{figure}[h]
        \centerline{\includegraphics[width=1\linewidth]{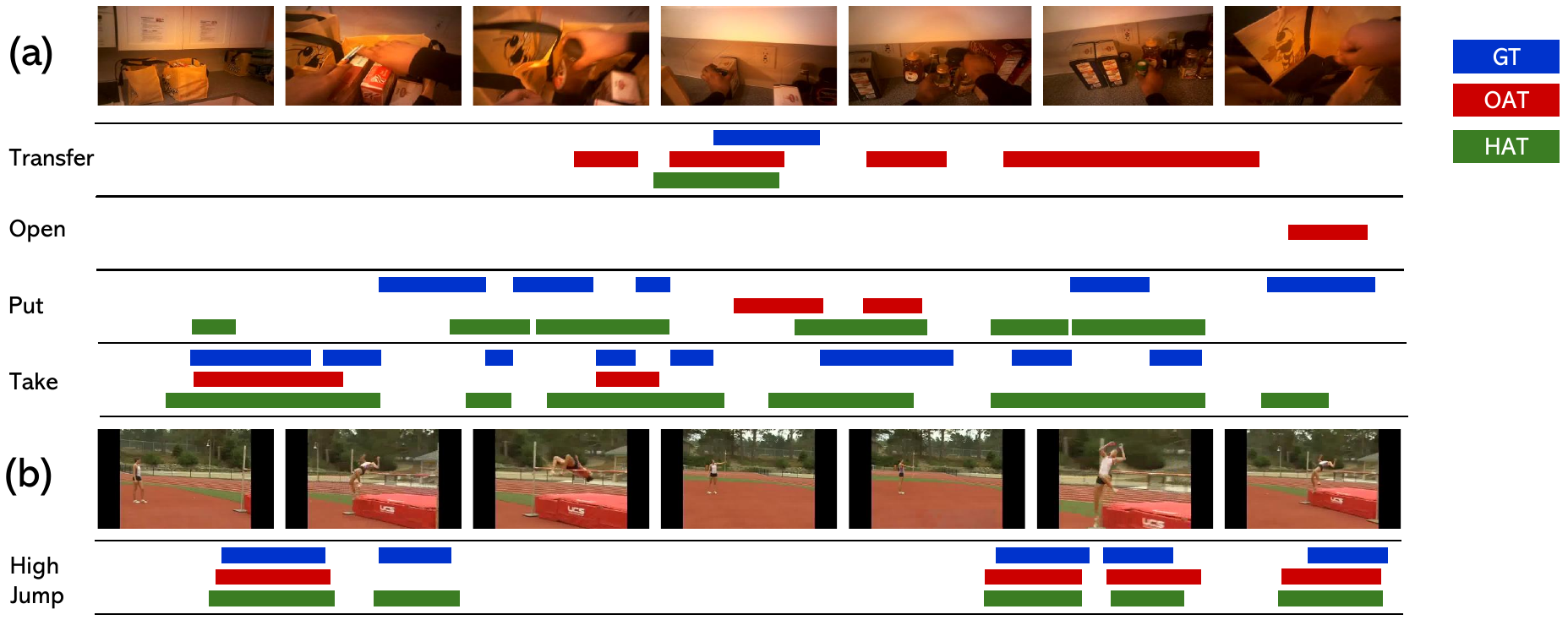}}
    \caption{Comparison of qualitative results of HAT with OAT \cite{kim2022sliding}. Blue is the ground truth, red is the result of OAT, and green is the result of HAT.} \label{qual1}
\end{figure}

In addition to standard quantitative evaluation methods, we conducted a manual qualitative analysis to determine the effectiveness of our proposed HAT model. Figure \ref{qual1}(a) presents a case from the procedural egocentric (PREGO) dataset EGTEA. Here, it is evident that the baseline model predicted a significantly higher number of false positives and missed action instances, failing to capture the broader context that our model provides. In Figure \ref{qual1}(b), we examine an example from the non-PREGO dataset THUMOS. In this scenario, the performance of the baseline and our model are nearly identical, with our model showing slight improvements in certain aspects. This qualitative assessment corroborates the findings from our previous quantitative analysis.

\subsection{Model Analysis}
In this section, we delve into the intricacies of the model by conducting ablation studies on the EGTEA dataset (split-1). These studies are designed to assess the effectiveness of various components within the proposed framework. 

\subsubsection{Impact of the History Module}
In this experiment, we evaluated the effectiveness of the Future-Supervised History Module (FSHM) by sequentially eliminating its components and observing the effects on performance across various tIOU thresholds. The experimental results are summarized in Table \ref{Tab:3}. 

\begin{table}
\tablefont{2.0mm} 
\begin{singlespace}
\caption{Impact of  history module on EGTEA dataset(Split-1). FSHM $=$ Future-Supervised History Module, AH $=$ Action Ancticipation Head, FDHR $=$ Future-Driven History Refinement.}
\label{Tab:3}
\centering
\begin{tabular}{l c c c c c c} 
 \toprule
  Model &  0.1 & 0.2 & 0.3 & 0.4 & 0.5 & Avg. \\ [0.5ex] 
 \hline
    w/ FSHM (Ours) & \textbf{27.3} & \textbf{25.3} & \textbf{21.6} & \textbf{16.9} & \textbf{12.8} & \textbf{20.8}
    \\
    w/o AH & 25.8 & 24.0 & 20.8 & \textbf{16.9} & 11.5 & 19.8
    \\
    w/o FDHR & 26.1 & 24.1 & 20.8 & 16.8 & 12.0 & 20.0
    \\
     w/o AH \& FDHR & 24.7 & 23.0 & 20.1 & 16.2 & 11.7 & 19.2 \\
     w/o FSHM & 24.2 & 22.7 & 20.5 & 16.3 & 10.9 & 18.9 \\
\bottomrule
\end{tabular}
\end{singlespace}
\end{table}

Removing the action anticipation head (`w/o AH'), which leaves only the history compressor and future-driven history refinement active, results in a notable performance decline across most tIOU thresholds. This indicates the critical role of the action anticipation head in calibrating the attention mechanism of the history compressor. In the scenario `w/o FDHR,' where the Future-Driven History Refinement component is omitted, there is a less severe performance decline,  showing that while important, FDHR's impact is slightly less than that of the action anticipation head. When both the action anticipation head and the future-driven history refinement are removed (`w/o AH \& FDHR'), leaving only the history compressor, there is an even more noticeable performance decrease, though still not as drastic as removing the entire module. Finally, completely eliminating the FSHM (`w/o FSHM') leads to a significant drop in performance across all tIOU thresholds, typically by about 2-3\%, which underlines the effectiveness of the proposed future-supervised history component in our framework.

\subsubsection{Comparison with other SOTA History Modules}
We conducted an analysis to assess the effectiveness of our history module compared to other state-of-the-art (SOTA) history processing methods. Since this is the first-ever exploration of historical context processing specifically for OnTAL, and no prior methods have been designed for OnTAL,  we opted to compare it with two
\begin{table}
\tablefont{2.0mm} 
\singlespacing
\centering
\caption{Comparison of our history processing method against other SOTA on EGTEA Dataset (Split-1).}
\label{Tab:4}
\begin{tabular}{l c c c c c c} 
 \toprule
  Model &  0.1 & 0.2 & 0.3 & 0.4 & 0.5 & Avg. \\ [0.5ex] 
 \hline
    No History & 24.2 & 22.7 & 20.5 & 16.3 & 10.9 & 18.9
    \\
    LSTR \cite{xu2021long} & 25.3 & 23.3 & 20.4 & 16.4 & 12.0 & 19.5
    \\
    GHU \cite{chen2022gatehub} & 26.2 & 24.0 & 21.0 & 16.6 & 11.7 & 19.9
    \\
     FSHM (Ours) & \textbf{27.3} & \textbf{25.3} & \textbf{21.6} & \textbf{16.9} & \textbf{12.8} & \textbf{20.8} \\
\bottomrule
\end{tabular}
\end{table}
  SOTA history modules developed for OAD tasks: GateHUB and LSTR. Table \ref{Tab:4} presents a comparison of our proposed future-supervised history module against alternative methods for processing historical data. Unlike the history module in LSTR, 
which employs a two-stage history compression method, our approach utilizes a single-stage
compression combined with an action anticipation head and future-driven history 
 refinement. The results indicate that our method performs better in the specific context of the OnTAL task.
 In GateHUB's history module, a one-stage history compression method is utilized, similar to ours, but it incorporates a gating mechanism to adjust the focus during the compression process, aiming to filter out irrelevant and background frames. However, our results demonstrate the superior performance of our approach, highlighting the effectiveness of our action anticipation guided attention calibration compared to their gating mechanism, especially within the context of the PREGO OnTAL task. Consequently, we can assert that our future-supervised history module is more effective for OnTAL applications than other SOTA history processing methods designed for OAD.

\subsubsection{Impact of the Adaptive Focal Loss}

In Table \ref{Tab:5}, we compare our introduced adaptive focal loss function with traditional cross-entropy, standard focal loss (with a 
constant  focal factor for all classes), and the background suppression focal (BSF) loss as utilized in GateHub (with distinct fixed focal factors for 
foreground and background). Our findings show that cross-entropy, regular focal loss, and 
BSF loss all result in lower mAP across various tIOUs when compared to our adaptive focal loss function. This underscores the effectiveness
 of applying adaptive focusing factors to the foreground classes based on the gradient
\begin{table}
\singlespacing
  \centering
  \caption{Comparison of different loss functions on EGTEA dataset (Split-1). }
\label{Tab:5}
\begin{tabular}{l c c c c c c} 
 \toprule
  Loss &  0.1 & 0.2 & 0.3 & 0.4 & 0.5 & Avg. \\ [0.5ex] 
 \hline
    Adaptive Focal Loss (Ours) & \textbf{27.3} & \textbf{25.3} & \textbf{21.6} & \textbf{16.9} & \textbf{12.8} & \textbf{20.8}
    \\
    BG-Sup. Focal Loss \cite{chen2022gatehub} & 27.2 & 24.4 & 21.1 & 16.7 & 11.5 & 20.2
    \\
    Regular Focal Loss \cite{lin2017focal} & 26.6 & 24.1 & 21.3 & 15.9 & 11.8 & 19.9
    \\
     Cross-Entropy Loss \cite{aldrich1997ra} & 25.1 & 22.9 & 20.2 & 16.6 & 12.1 & 19.4 \\
\bottomrule
\end{tabular}
\end{table}
  ratio of positive-to-negative samples during training, in addition to applying a distinct focal factor for the background class. This approach effectively addresses the issue of class imbalance between foreground and background, as well as among the foreground classes themselves. Whereas the BSF loss previously addressed only the foreground-background class imbalance through two separate focusing factors, it failed to address the imbalance among the foreground classes. The superior performance of our adaptive focal loss demonstrates its capacity to more effectively focus on the less confident, imbalanced classes during training, leading to improved outcomes.

\subsubsection{Qualitative Analysis of the History Module}
\begin{figure}[t!]
	\centerline{\includegraphics[width=0.9\linewidth]{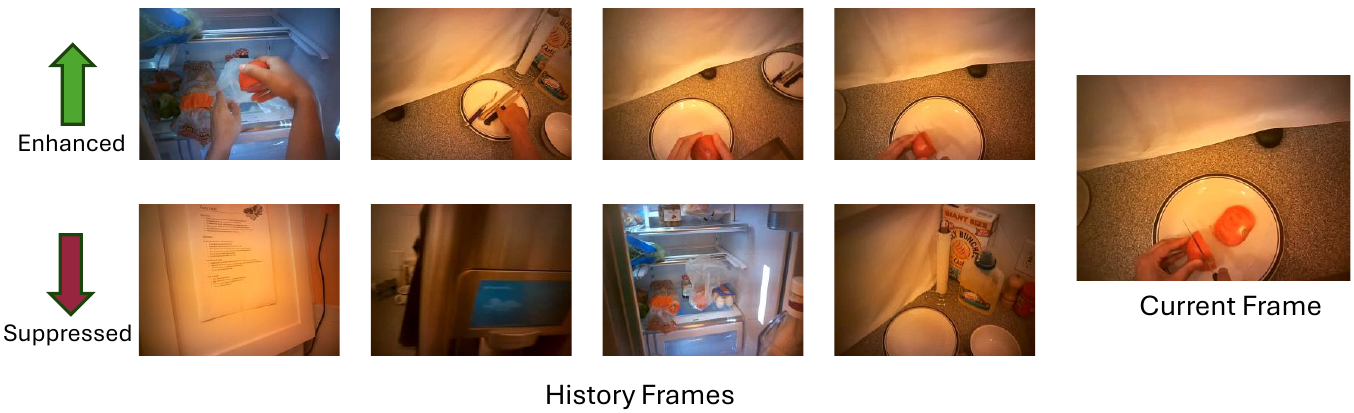}}
    \caption{Frame attention analysis for an ``cutting'' a tomato example from EGTEA dataset. The frames with related actions such as ``taking'' the tomato from the freezer, ``taking'' a knife, ``putting'' the tomato on a plate and the initial ``cutting'' action receive more attention (upper row).} \label{qual2}
\end{figure}

To assess the effectiveness of our future-supervised history module beyond quantitative metrics, we conducted a qualitative analysis by examining the attention weights. This evaluation aims to determine if the history compressor within the module can successfully downplay irrelevant frames while highlighting those crucial to the current action. A representative example in Figure \ref{qual2} demonstrates this by showcasing the subject performing a ``cutting'' action on a tomato. In this context, we divided the historical frames into two groups: those receiving higher attention than the median, and those receiving lower. The frames receiving more focus depict pertinent activities such as ``taking'' the tomato from the freezer, ``taking'' a knife, ``putting'' the tomato on a plate and the initial ``cutting'' action, illustrating the compressor’s ability to highlight relevant action-related history. Conversely, the less emphasized frames mostly include irrelevant egocentric movements and unrelated background details, underscoring the module's capability to filter out irrelevant information to the ongoing action.
 
\section{Conclusions and Future Work}
In conclusion, we present a novel approach to Online Temporal Action Localization (OnTAL) by introducing a History-Augmented Anchor Transformer, which innovatively combines long-term historical information with anchor-level features. We designed a unique history processing method, incorporating action anticipation-guided history compression and future context-driven alignment, to enhance instance-level action prediction performance. Additionally, We introduced an adaptive focal loss function that addresses the class imbalance issue between foreground and background, as well as among different foreground classes. Our comprehensive analysis across four datasets demonstrates the effectiveness of our approach on the OnTAL task, particularly highlighting the influence of our history processing and integration in procedural egocentric (PREGO) contexts.

Our framework, while capable of utilizing long-term history, is restricted by a fixed history length. This creates a challenge as determining the optimal length for this history involves extensive manual adjustments. Furthermore, different instances within the same use case may require varying lengths of history—one may need a shorter span, while another may require a longer one. Although our system can filter out irrelevant information and focus on pertinent details within this fixed span, it lacks the ability to dynamically modify the history span. It would be ideal if the history component could automatically adjust its length to capture the long-term history and allocate attention to specific frames as needed, similar to the adaptive attention span techniques used in language models \cite{sukhbaatar2019adaptive}. This adjustment would not only enable more effective long-term context capture but also reduce overall computational demands.

\section*{Acknowledgments}
This work is supported by the U.S. National Science Foundation (NSF) Grants No. FW-HTF-2128743, IIS-2302838, IIS-1814631, and CNS-2038493, ONR grant N00014-21-1-2431 from NCI, and U.S. Department of Homeland Security grant 22STESE00001-03-02. The views and conclusions contained in this document are those of the authors and should not be interpreted as necessarily representing the official policies, either expressed or implied, of the U.S. Department of Homeland Security, the NSF and the ONR.
%
%
\bibliographystyle{splncs04}
\bibliography{main}

\pagebreak
\appendix
\section*{Appendix}
\section{Method}
\label{appx:method}
\subsection{Anchor Feature Generation}
The process of generating anchor features from short-term windows involves two main stages. Initially, a transformer encoder, composed of \(N_e\) sequential blocks, is utilized to convert local attributes into a temporal contextual feature. Subsequently, this temporal feature is supplied to a transformer decoder, which incorporates learnable anchor tokens to synthesize features for each distinct anchor, similar to the approach in \cite{kim2022sliding}.

Each block within the transformer encoder, denoted as $e$, comprises a multihead self-attention mechanism and a feedforward network (FFN). The operational mechanism of a single encoder block can be represented as follows:

\begin{align}
e(X) = \text{Norm}(X_\text{attn} + \text{FFN}(X_\text{attn})), \\
X_\text{attn} = \text{Norm}(X + \text{SelfAttn}(X)),
\end{align}
where \(\text{FFN}\) stands for the feedforward network, and \(\text{SelfAttn}\) denotes the multi-head self-attention layer. As the model encompasses $N_e$ encoder blocks, the final output is expressed as \( S_\text{enc} = e_{N_e} \circ \ldots \circ e_1 (S) \in \mathbb{R}^{L_s \times D} \), where \(L_s\) is the length of the short-term temporal context, and D represents the dimensionality of the features.

In the proposed model, each new video frame triggers the generation of action proposals, capitalizing on predefined anchor templates reflecting true event boundaries. Traditionally, this required the manual normalization of anchor segments of disparate lengths into a uniform feature format through different pooling strategies. However, following \cite{kim2022sliding}, we simplify this by directly deriving anchor representations, thereby eliminating the conventional requirements for manual size adjustment and reshaping. This is achieved by transforming the output sequence from the encoder, \(S_\text{enc}\), which has a length of \(L_s\), into a new sequence of representations of size M, using a standard transformer decoder.

The transformer decoder receives two types of input: a learnable anchor query \(Q_\text{anc} \in \mathbb{R}^{M \times D}\), and the encoded short-term temporal context, \(S_\text{enc}\). As it progresses through \(N_d\) decoder blocks, each containing multi-head cross-attention and feedforward network (FFN) layers, it produces an output $A$ \( \in \mathbb{R}^{M \times D}\). 

\subsection{Prediction Module and Online Post-Processing}
Following \cite{kim2022sliding}, the prediction module consists of two distinct components: an action classifier and a boundary regressor. Both components use the same processed anchor feature representation \(A\) for their input. The function of the action classifier is to identify the specific action taking place within a given anchor segment and to decide whether this segment contains segments of non-action (such as background scenes). Conversely, the regressor is designed to compute two key measurements: the distance or offset between the end of the target action and the anchor segment's end, and the ratio of the duration of the target action to the total duration of the anchor segment. Each module consists of a neural network with two layers that produce distinct sets of output values. For the action classification part, this output is a set of classification scores, \( z_c \), in the space \(\mathbb{R}^{M \times (C+1)}\), which includes predictions for different action categories as well as the background category. Meanwhile, the boundary regression part generates a pair of scores, \(\{z_o, z_l\}\), in the space \(\mathbb{R}^{M \times 2}\), that represents the estimated offset and length ratio, respectively.

In addressing the issue of removing repetitive action proposals in video streams without relying on future data, we follow a online post-processing method described in the referenced work \cite{kim2022sliding}. This method involves a neural network that selects the most appropriate action proposal at each time step, akin to Non-Maximum Suppression (NMS). The approach involves generating \(M\) proposals for each frame, eliminating those associated with the background, and executing NMS on the outcomes of each single frame. The confidence scores are maintained in a history buffer \(q_h\) with dimensions \([0, 1]^{Ls\times C}\), which affects the network's probability of NMS selection within the range \([0, 1]^C\). If the probability surpasses the preset threshold \(\theta_s\), the proposal is included in the final set of instances \(\Psi\), and any following proposals that are similar are disregarded. The training process requires constructing input tables based on the main model's confidence scores and the actual labels derived from NMS outcomes, with the network being optimized via binary cross-entropy loss. 

\subsection{Training Strategy and Loss Function}
\label{appx:training}
\subsubsection{Adaptive Focal Loss}
In the video action understanding datasets, a common issue is the imbalance between action samples and background samples, with the latter usually dominating. The focal loss was used as a solution to this imbalance by assigning varying levels of emphasis to different instances based on their classification confidence \cite{chen2022gatehub}. Specifically, it differentiates between hard-to-classify action instances and background instances by applying distinct focal factors, thereby addressing the imbalance between action and background classes.

However, this approach does not address the potential imbalance among different action classes themselves. In response, we introduce a modified version of the focal loss, Adaptive Focal Loss (AFL), tailored for OnTAL. This new loss function dynamically adjusts the focal factor for each action class during training, informed by the gradients, to better manage class imbalance across various action categories. The Adaptive Focal Loss is formulated as follows:

\begin{align}
AFL(p_i, y_i) = \sum_{j=0}^{C} -y_i^j (1 - p_i^j)^{\lambda^j} \log(p_i^j),
\end{align}
where \( p_i \) is the predicted probability for the \(i^{th}\) anchor, \( y_i \) is the corresponding ground truth label, and \( \lambda^j \) is the class-specific focal factor for the \(j^{th}\) action category. This class-specific focal factor, \( \lambda^j \), is defined as:

\begin{align}
\lambda^j = \begin{cases}
      \lambda_b, & \text{for } j = C \\
      \lambda_b + \lambda^j_f, & \text{otherwise}
    \end{cases}, \\
     \lambda^j_f = s(1 - r^j) ,
\end{align}
where \( \lambda_b \) is the base focal factor, and \( \lambda^j_f\) is an adaptive component for the foreground classes, with \( r^j \) being the mapped ratio of accumulated gradients for positive versus negative samples for the \(j^{th}\) category. The scaling factor \( s \) adjusts the influence of \( \lambda^j_f \). The value of \(r^j\) is adjusted so that it ranges only from 0 to 1. The computation of \(r^j\) proceeds as follows:

\begin{align}
r^{j} = f(g^{j}), \\
g^{j} = \frac{g_\text{pos}^{j}}{g_\text{neg}^{j}}, \\
f(x) = \frac{1}{1 + e^{-(n(x)-\mu)}},
\end{align}
where the terms \(g_\text{pos}^{j}\) and \(g_\text{neg}^{j}\) represent the gradients calculated from positive and negative samples, respectively, as used in \cite{li2022equalized}. \(f(x)\) is the mapping function and \(n(x)\) is the min-max normalization operation \cite{patro2015normalization}. \(\mu\) is the mean of all the normalized gradient ratios in the batch. The value of \( r^j \) approaches 1 for categories with less imbalance, thereby reducing the additional focal impact, and approaches 0 for categories with greater imbalance, thereby increasing the additional focal impact. 

When setting values for equation \textcolor{red}{4}, we empirically found that setting the value of \(s = 2 \lambda_b\) works well. For instance, in an average case (i.e., the relative imbalance is closer to the batch’s mean), if we want to set \(\lambda_j = 0.025\) for the background class (\(j=C\)) and \(\lambda_j = 0.05\) for all other (foreground) classes (\(j \neq C\)), we should set \(\lambda_b = 0.025\) and \(s = 0.05\). With this setup, when the normalized gradient ratio \(n(g^j)\) for a class is around the batch's mean, the adjusted gradient ratio \(r^j\) will be approximately 0.5. This makes the \(\lambda_j\) value for foreground classes about 0.05, as intended. If a foreground class is significantly more or less imbalanced compared to the mean, \(\lambda_j\) will adjust accordingly, becoming either larger or smaller than 0.05.

\subsubsection{Final Loss} 
In the training process, we guide the output of the prediction module using pre-defined anchors and the output of the action anticipation head using a short-term window. Specifically, for the prediction module, we establish anchors whose end points are aligned with the most recent input frame, creating M different lengths of anchors (such as \(\{L_s/8, L_s/4, L_s/2, L_s\}\)). We then assess the overlap between these anchors and the actual ground truth instances by calculating the Intersection-over-Union (IoU) for each pair. An anchor is considered a match to a ground truth if their IoU exceeds the defined matching threshold \(\theta_m\). Once matched, we compute the losses for the prediction module based on these associations. Separately, for the action anticipation head, we assess its ability to predict actions occurring within the short-term window and compute its loss based on this assessment. The methodology for calculating the losses in our model is detailed below:

\begin{align}
    \mathcal{L}_c = \sum_{i=1}^{K} \text{AFL}(\text{softmax}(z_{c,i}), y_{c,i}), \\
    \mathcal{L}_o = \sum_{i=1}^{K} \text{L1}\left(z_{o,i} - \frac{y_{e,i} - a_{e,i}}{a_{l,i}}\right), \\
    \mathcal{L}_l = \sum_{i=1}^{K} \text{L1}\left(z_{l,i} - \log\frac{y_{l,i}}{a_{l,i}}\right),\\
    \mathcal{L}_a = \text{AFL}(\text{softmax}(z_{c,i}), y_{c,i}),\\
    \mathcal{L} = \alpha\mathcal{L}_c + \beta(\mathcal{L}_o + \mathcal{L}_l)+\gamma\mathcal{L}_a,
\end{align}
Here, we utilize adaptive focal loss (AFL) for classifying anchor segments and apply L1 loss for regression, where \(y_c\), \(y_e\), \(y_l\), \(a_e\), and \(a_l\) represent actual class, true action end, true action length, anchor segment end, and anchor segment length, respectively. The total loss is determined by combining various losses using weighted coefficients. For prediction module, anchors meeting specific criteria are marked as foreground and undergo boundary regression; otherwise, they are marked as background, skipping regression. The overall loss is calculated combining different loss with weighted co-efficients \(\alpha, \beta, \text{and } \gamma\). Proposals are generated by fixing anchor ends to the latest frame, ensuring early actions are classified as background, while middle and end actions, aligned with true actions, allow for accurate boundary predictions.

During inference, our framework generates \(M\) action proposals \(\{(s_i, e_i, c_i)\}_{i=1}^M\) for every timestep. These proposals are defined as follows:
\begin{align}
c_i = \mathrm{softmax}(\text{z}_{c,i}), \\
e_i = a_{o,i} + a_{l,i} \cdot \text{z}_{e,i}, \\
s_i = e_i - a_{l,i} \cdot \exp(\text{z}_{l,i}).
\end{align}

Note that in this inference process, the action anticipation head from the history module is not used and should be ignored.

\section{Experiments}

\subsection{Dataset}

Although the primary details about the dataset have already been presented in the main paper, here are some further insights and specifics regarding the datasets provided.

\subsubsection{EGTEA\cite{li2018eye}} We extracted features using the I3D model \cite{i3d}, which was pretrained on the Kinetics dataset. These features were processed at their original 24 FPS frame rate with a stride of 12, leading to two feature vectors being generated every second. We adopted a four-fold cross-validation approach for training and evaluation, similar to \cite{huang2020improving}. For evaluation purposes, we employed the metric mAP@[0.1:0.1:0.5] and also reported the average mAP score. The action annotations in this dataset were solely based on verb (action) annotations.

\subsubsection{EPIC-Kitchens-100\cite{damen2022rescaling}} Feature extraction for this dataset was performed using the SlowFast model \cite{feichtenhofer2019slowfast}, which had been pretrained on the EPIC Kitchens 100 training set specifically for action classification. This process involved the use of 32-frame clips at a 30 FPS frame rate and a stride of 16 frames, resulting in one feature vector approximately every 0.5333 seconds. Our model underwent training on the training set and was subsequently evaluated on the validation set. The evaluation metric utilized was mAP@[0.1:0.1:0.5], with the average mAP being reported in alignment with previous standards \cite{zhang2022actionformer, damen2022rescaling}. Actions in this dataset are identified by a combination of a verb (action) and a noun (object), though our analysis solely focused on verb (action) annotations to maintain consistency with other OnTAL research.

\subsubsection{THUMOS'14\cite{idrees2017thumos}} This dataset comprises 413 untrimmed videos spanning 20 action categories, divided into validation (200 videos) and test (213 videos) sets. Consistent with established practices, we used the validation set for training purposes and reported our findings on the test set. Feature extraction was conducted using the two-stream TSN model \cite{wang2016temporal}, pretrained on the Kinetics dataset \cite{i3d}, in accordance with prior methodologies. The evaluation metric applied was mAP@[0.3:0.1:0.7].

\subsubsection{MUSES\cite{liu2021multi}} For MUSES, we utilized the I3D \cite{i3d} features that were officially provided. The training and testing phases were conducted using the original train-test subsets designated for the dataset. The evaluation followed a similar framework to that of THUMOS’14, using mAP@[0.3:0.1:0.7] as the metric and reporting the average mAP.

\subsection{Additional Model Analysis}

\subsubsection{History Length}
Although adjusting the history length in our model is not a principal focus of our study, we conducted an analysis to illustrate the impact of history length on model performance using the EGTEA dataset (split-1), with findings presented in Table \ref{Tab:6}. The analysis reveals an absence of definitive trends in performance across various history lengths; however, it indicates that increasing the history length tends to enhance performance up to a certain threshold. For the EGTEA dataset, an optimal history length of 24 seconds was identified.

\begin{table}
  \centering
\caption{History length analysis of HAT on EGTEA dataset (split-1) across various tIOU thresholds. }
\label{Tab:6}
  \begin{tabular}{c c c c c c c} 
 \toprule
  History Length $L_h$ (sec.) &  0.1 & 0.2 & 0.3 & 0.4 & 0.5 & Avg. \\ [0.5ex] 
 \hline
     0 & 24.2 & 22.7 & 20.5 & 16.3 & 10.9 & 18.9 \\
     8 & 25.0 & 23.1 & 19.7 & 16.1 & 12.0 & 19.2 \\
     16 & 26.5 & 24.5 & 21.6 & 16.7 & 11.7 & 20.2 \\
     24 & 27.3 & \textbf{25.3} & 21.6 & \textbf{16.9} & \textbf{12.8} & \textbf{20.8} \\
     32 & 27.0 & 24.5 & 21.1 & 16.6 & 11.7 & 20.2 \\
     40 & \textbf{27.5} & 25.2 & \
     \textbf{22.0} & 16.6 & 11.8 & 20.6 \\
    
\bottomrule
\end{tabular}

\end{table}

\begin{figure}
	\centerline{\includegraphics[width=0.7\linewidth]{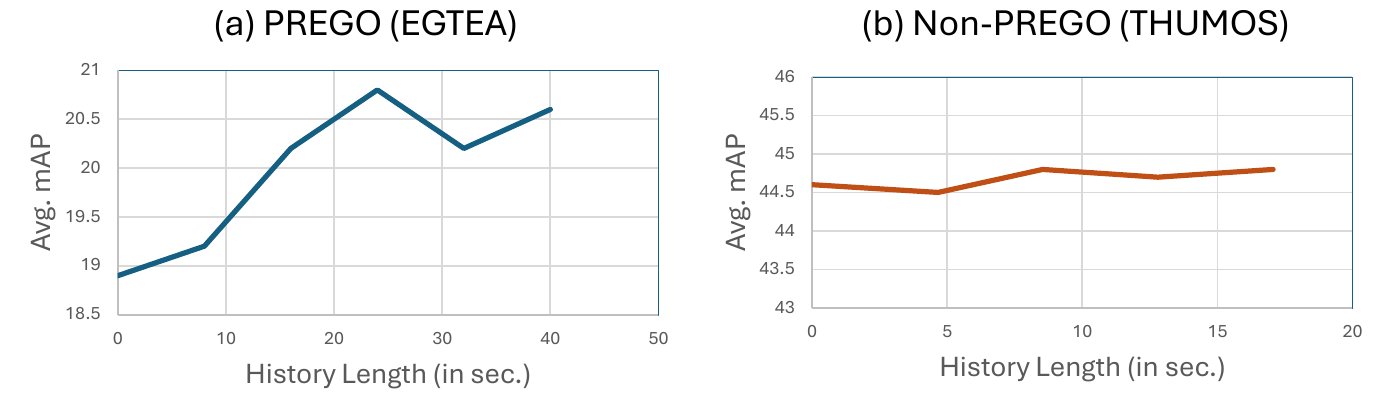}}
    \caption{Impact of history length comparison} \label{len}
    \vspace{-4mm}
\end{figure}

Furthermore, we compared the impact of history length between a procedural egocentric (PREGO) dataset (i.e., EGTEA) and a non-PREGO dataset (i.e., THUMOS), as shown in Figure \ref{len}. The results indicate that in the PREGO scenario, increasing the history length positively impacts performance until it reaches a saturation point. Conversely, in the non-PREGO scenario, increasing the history length does not significantly affect performance.

\subsubsection{History Integration Method}
Prior research \cite{xu2021long} shows that the transformer decoder is the most effective method for combining historical features with current features. Nevertheless, we conducted an analysis to compare various approaches, considering our unique context for integrating historical and anchor features. Table \ref{Tab:7} presents the comparison among these methods, demonstrating that the transformer decoder outperforms the alternatives.

\begin{table}
  \centering
\caption{Comparison of History Integration Methods in History-Augmented Anchor Module on EGTEA (Split-1). }

\begin{tabular}{l c c c c c c} 
 \toprule
  Method &  0.1 & 0.2 & 0.3 & 0.4 & 0.5 & Avg. \\ [0.5ex] 
 \hline

    w/ Average Pooling & 24.6 & 23.0 & 20.4 & 16.5 & 11.7 & 19.3
    \\
    w/ Concatenation  & 25.3 & 23.0 & 20.3 & 16.7 & 12.1 & 19.5 \\
    \footnotesize{w/ Temporal NLB (Cross-Attn.) \cite{xu2021long}} & 26.8 & 24.7 & 21.2 & \textbf{17.0} & 12.0 & 20.3
    \\
    \textbf{Ours} (w/ Trans. Decoder) & \textbf{27.3} & \textbf{25.3} & \textbf{21.6} & 16.9 & \textbf{12.8} & \textbf{20.8}
    \\
\bottomrule
\end{tabular}

\label{Tab:7}
\end{table}

\subsubsection{Early Detected Time}
Similar to the baseline OAT model \cite{kim2022sliding}, our proposed HAT model is also capable of predicting an action instance before the action is completed. While we made no specific efforts to enhance the responsiveness of our model, we conducted an evaluation to determine if incorporating long-term historical data impacts its responsiveness. To assess this, we used the Average Early Detection Time (AEDT) metric \cite{kim2022sliding} for comparison with the baseline. 
\begin{table}[H]
  \centering
  \caption{Average Early Detected Time (AEDT) comparison of our HAT model with the baseline OAT model across various tIOU thresholds. }
\label{Tab:8}

  \begin{tabular}{ l c c c c c c} 
 \toprule
   Model &  0.3 & 0.4 & 0.5 & 0.6 & 0.7 & Avg. \\ [0.5ex] 
 \hline

     OAT\cite{kim2022sliding} & \textbf{-0.96} & \textbf{-1.01} & \textbf{-1.02} & -1.00 & \textbf{-1.03} & \textbf{-1.01} \\
     HAT (Ours) & -0.83 & -0.90 & -1.01 & \textbf{-1.06} & \textbf{-1.03} & -0.97 \\
    
\bottomrule
\end{tabular}

\end{table}
According to the results presented in Table \ref{Tab:8}, our model exhibits performance on par with the baseline in terms of early detection. On average, both models are able to detect actions approximately 1 second before they conclude. This suggests that the integration of long-term history does not significantly affect the model's promptness.

\end{document}